\documentclass[11pt]{article}
\usepackage{coling2020}
\usepackage{times}
\usepackage{url}
\usepackage{latexsym}

\usepackage{graphicx}
\usepackage{booktabs}
\usepackage{subfigure}
\usepackage[table,xcdraw]{xcolor}
\usepackage{amsmath}
\usepackage{wrapfig}
\usepackage[normalem]{ulem}
\useunder{\uline}{\ul}{}
\usepackage{multirow}
\usepackage[title]{appendix}

\usepackage{floatrow}
\newfloatcommand{capbtabbox}{table}[][\FBwidth]

\usepackage[pagebackref=true,breaklinks=true,colorlinks=false,bookmarks=false]{hyperref}

\usepackage{breakurl}

\colingfinalcopy 

\title{Catching Attention with Automatic Pull Quote Selection}

\author{Tanner Bohn \\
  Western University \\
  London, ON, Canada \\
  {\tt tbohn@uwo.ca} \\\And
  Charles X. Ling \\
  Western University \\
  London, ON, Canada \\
  {\tt charles.ling@uwo.ca} \\}

\date{}

\begin{document}
\maketitle
\begin{abstract}
To advance understanding on how to engage readers, we advocate the novel task of automatic pull quote selection. Pull quotes are a component of articles specifically designed to catch the attention of readers with spans of text selected from the article and given more salient presentation. This task differs from related tasks such as summarization and clickbait identification by several aspects. We establish a spectrum of baseline approaches to the task, ranging from handcrafted features to a neural mixture-of-experts to cross-task models. By examining the contributions of individual features and embedding dimensions from these models, we uncover unexpected properties of pull quotes to help answer the important question of what engages readers. Human evaluation also supports the uniqueness of this task and the suitability of our selection models. The benefits of exploring this problem further are clear: pull quotes increase enjoyment and readability, shape reader perceptions, and facilitate learning. Code to reproduce this work is available at \url{https://github.com/tannerbohn/AutomaticPullQuoteSelection}.

\end{abstract}

%
%
\blfootnote{
    %
    %
    
    \hspace{-0.65cm}  
    This work is licensed under a Creative Commons 
    Attribution 4.0 International License.
    License details:
    \url{http://creativecommons.org/licenses/by/4.0/}.
}

\section{Introduction}


\begin{wrapfigure}{r}{0.4\textwidth}
	\centering
	\includegraphics[width=1\columnwidth]{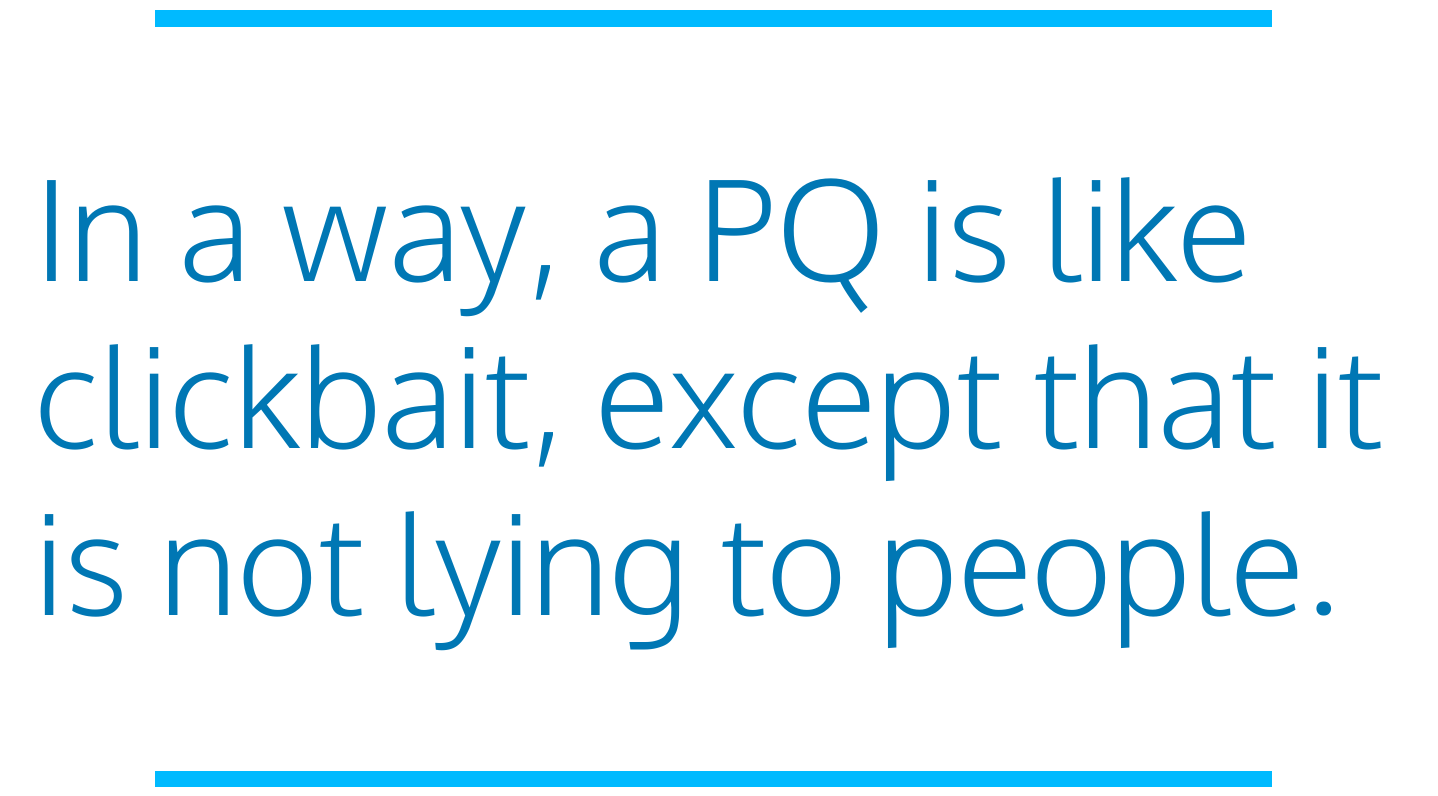}
	\caption{A pull quote from this paper chosen with the help of our best performing model (see Section~\ref{sec:results_embeddings}).}
	\label{fig:paper_pq}
\end{wrapfigure}

Discovering what keeps readers engaged is an important problem.
We thus propose the novel task of automatic pull quote (PQ) selection accompanied with a new dataset and insightful analysis of several motivated baselines. PQs are graphical elements of articles with thought provoking spans of text pulled from an article by a writer or copy editor and presented on the page in a more salient manner \cite{french2018indesign}, such as in Figure~\ref{fig:paper_pq}.

PQs serve many purposes. They provide temptation (with unusual or intriguing phrases, they make strong entrypoints for a browsing reader), emphasis (by reinforcing particular aspects of the article), and improve overall visual balance and excitement \cite{stovall1997infographics,holmes2015subediting}. PQ frequency in reading material is also significantly related to information recall and student ratings of enjoyment, readability, and attractiveness \cite{wanta1994young,wanta1994information}.


The problem of automatically selecting PQs is related to the previously studied tasks of headline success prediction \cite{piotrkowicz2017headlines,lamprinidis2018predicting}, clickbait identification \cite{potthast2016clickbait,chakraborty2016stop,venneti2018curiosity}, as well as key phrase extraction \cite{hasan2014automatic} and document summarization \cite{nenkova2012survey}. However, in Sections~\ref{sec:results_cross_task} and \ref{sec:human_eval} we provide experimental evidence that performing well on these previous tasks does not translate to performing well at PQ selection. Each of these types of text has a different function in the context of engaging a reader. The title tells the reader what the article is about and sets the tone. Clickbait makes unwarranted enticing promises of what the article is about. Key phrases and summaries help the reader decide whether the topic is of interest. And PQs provide \textit{specific} intriguing entrypoints for the reader or can \textit{maintain} interest once reading has begun by providing glimpses of interesting things to come. With their unique qualities, we believe PQs satisfy important roles missed by these popular existing tasks.


\begin{figure}[ht]
	\centering
	\includegraphics[width=1\columnwidth]{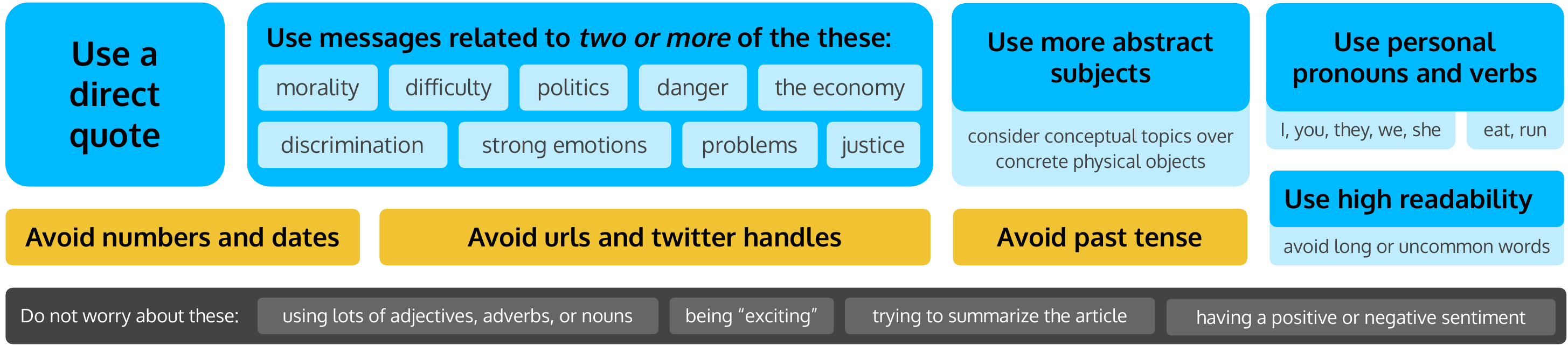}
	\caption{Factors suggested by our results to be important (and unimportant) in creating pull quotes.}
	\label{fig:advice}
\end{figure}


In this work we define PQ selection as a sentence classification task and create a dataset of articles and their expert-selected PQs from a variety of news sources.
We establish a number of approaches with which to solve and gain insight into this task: (1) handcrafted features, (2) n-gram encodings, (3) Sentence-BERT (SBERT) \cite{reimers2019sentence} embeddings combined with a progression of neural architectures, and (4) cross-task models. 
Via each of these model groups, we uncover interesting patterns (summarized in Figure~\ref{fig:advice}). For example, among handcrafted features, sentiment and arousal are surprisingly uninformative features, overshadowed by presence of quotation marks and reading difficulty. Analysing individual SBERT embedding dimensions also helps understand the particular themes that make for a good PQ. We also find that combining SBERT sentence and document embeddings in a mixture-of-experts manner provide the best performance at PQ selection. The suitability of our models at PQ selection is also supported via human evaluation.


The main contributions are:

\begin{enumerate}
    
    \item We describe several motivated approaches for the new task of PQ selection, including a mixture-of-experts approach to combine sentence and document embeddings (Section~\ref{sec:models}).
    
    \item We construct a dataset for training and evaluation of automatic PQ selection (Section~\ref{sec:setup}). 
    
    \item We inspect the performance of our approaches to gain a deeper understanding of PQs, their relation to other tasks, and what engages readers (Section~\ref{sec:results}). Figure.~\ref{fig:advice} summarizes these findings.

\end{enumerate}

\section{Related Work}
\label{sec:related}



In this section, we look at three natural language processing tasks related to PQ selection: (1) headline quality prediction, (2) clickbait identification, and (3) summarization and keyphrase extraction. These topics motivate the cross-task models whose performance on PQ selection is reported in Section~\ref{sec:results_cross_task}.

\subsection{Headline Quality Prediction}

When a reader comes across a news article, the headline is often the first thing given a chance to catch their attention, thus predicting their success is a strongly motivated task. Once a reader decides to check out the article, it is up to the content (including PQs) to maintain their engagement. 



In \cite{piotrkowicz2017headlines}, the authors experimented with two sets of features: journalism-inspired (which aim to measure how news-worthy the topic itself is), and linguistic style features (reflecting properties such as length, readability, and parts-of-speech -- we consider such features here). They found that overall the simpler style features work better than the more complex journalism-inspired features at predicting social media popularity of news articles. The success of simple features is also reflected in \cite{lamprinidis2018predicting}, which proposed multi-task training of a recurrent neural network to not only predict headline popularity given pre-trained word embeddings, but also predict its topic and parts-of-speech tags. They found that while the multi-task learning helped, it performed only as well as a logistic regression model using character n-grams. 
Similar to these previous works, we also evaluate several expert-knowledge based features and n-grams, however, we expand upon this to include a larger variety of models and provide a more thorough inspection of performance to understand what engages readers.


\subsection{Clickbait Identification}

The detection of a certain type of headline -- clickbait -- is a recently popular task of study. Clickbait is a particularly catchy headline and form of false advertising used by news outlets which lure potential readers but often fail to meet expectations, leaving readers disappointed \cite{potthast2016clickbait}. Clickbait examples include ``You Won't Believe...'' or ``X Things You Should...''. We suspect that the task of distinguishing between clickbait and non-clickbait headlines is related to PQ selection because both tasks may rely on identifying the catchiness of a span of text. However, PQs attract your attention with content truly in the article. In a way, a PQ is like clickbait, except that it is not lying to people.

In \cite{venneti2018curiosity}, the authors found that measures of topic novelty (estimated using LDA) and surprise (based on word bi-gram frequency) were strong features for detecting clickbait. In our work however, we investigate the interesting topics themselves (Section~\ref{sec:results_embeddings}). A set of 215 handcrafted features were considered in \cite{potthast2016clickbait} including sentiment, length statistics, specific word occurrences, but the authors found that the most successful features were character and word n-grams. The strength of n-gram features at this task is also supported by \cite{chakraborty2016stop}. While we also demonstrate the surprising effectiveness of n-grams and consider a variety of handcrafted features for our particular task, we examine more advanced approaches that exhibit superior performance.

\subsection{Summarization and Keyphrase Extraction}


Document summarization and keyphrase extraction are two well-studied NLP tasks with the goals of capturing and conveying the main topics and key information discussed in a body of text \cite{turney1999learning,nenkova2012survey}. Keyphrase extraction is concerned with doing this at the level of individual phrases, while extractive document summarization (which is just one type of summarization \cite{nenkova2011automatic}) aims to do this at the sentence level. Approaches to summarization have roughly evolved from unsupervised extractive heuristic-based methods \cite{luhn1958automatic,mihalcea2004textrank,erkan2004lexrank,nenkova2005impact,haghighi2009exploring}, to supervised and often abstractive deep-learning approaches \cite{nallapati2016classify,nallapati2016abstractive,nallapati2017summarunner,zhang2019pegasus}. Approaches to keyphrase extraction fall into similar groups, with unsupervised approaches including \cite{tomokiyo2003language,mihalcea2004textrank,liu2009clustering}, and supervised approaches including \cite{turney1999learning,medelyan2009human,romary2010automatic}.

While summarization and keyphrase extraction are concerned with what is \textit{important} or representative in a document, we instead are interested in understanding what is \textit{engaging}. While these two concepts may seem very similar, in Sections~\ref{sec:results_cross_task} and \ref{sec:results_cross_task} we provide evidence of their difference by demonstrating that what makes for a good summary does not make for a good PQ.

\section{Models}
\label{sec:models}


We consider four groups of approaches for the PQ selection task:
(1) handcrafted features (Section~\ref{sec:models_hc}), 
(2) n-gram features (Section~\ref{sec:models_ngram}), 
(3) SBERT embeddings combined with a progression of neural architectures (Section~\ref{sec:models_embeddings}), and
(4) cross-task models (Section~\ref{sec:models_cross_task}).
As discussed further in Section~\ref{sec:setup}, these approaches aim to determine the probability that a given article sentence will be used for a PQ.

\subsection{Handcrafted Features}
\label{sec:models_hc}

Our handcrafted features can be loosely grouped into three categories: surface, parts-of-speech, and affect, each of which we will provide justification for. For the classifier we will use AdaBoost \cite{hastie2009multi} with a decision tree base estimator, as this was found to outperform simpler classifiers without requiring much hyperparameter tuning.

\subsubsection{Surface Features}
\label{sec:models_hc_surface}

    \begin{itemize}
        \item \textbf{Length}: We expect that writers have a preference to choose PQs which are concise. To measure length, we will use the total character length, as this more accurately reflects the space used by the text than the number of words.

        \item \textbf{Sentence position}: We consider the location of the sentence in the document (from 0 to 1). This is motivated by the finding in summarization that summary-suitable sentences tend to occur near the beginning \cite{braddock1974frequency} -- perhaps a similar trend exists for PQs.
        
        \item \textbf{Quotation marks}: We observe that PQs often contain content from direct quotations. As a feature, we thus include the count of opening and closing double quotation marks.

        \item \textbf{Readability}: Motivated by the assumption that writers will not purposefully choose difficult-to-read PQs, we consider two readability metric features:
(1) \textbf{Flesch Reading Ease}: This measure ($R_{Flesch}$) defines reading ease in terms of the number of words per sentence and the number of syllables per word \cite{flesch1979write}.
(2) \textbf{Difficult words}: This measure ($R_\mathit{difficult}$) is the percentage of unique words which are considered ``difficult'' (at least six characters long and not in a list of $\sim$3000 easy-to-understand words). See Appendix~\ref{sec:implementation} for details.

    \end{itemize}

\subsubsection{Part-of-Speech Features}



We include the word density of part-of-speech (POS) tags in a sentence as a feature. As suggested by \cite{piotrkowicz2017headlines} with respect to writing good headlines, we suspect that verb (VB) and adverb (RB) density will be informative. 
We also report results on the following: cardinal digit (CD), adjective (JJ), modal verb (MD), singular noun (NN), proper noun (NNP), personal pronoun (PRP).


\subsubsection{Affect Features}


Events or images that are shocking, filled with emotion, or otherwise exciting will attract attention \cite{schupp2007selective}.
However, this does not necessarily mean that text describing these things will catch reader interest as reliably \cite{aquino2007attention}. 
To determine how predictive sentence affect properties are of PQ suitability, we include the following features:


\textbf{Positive sentiment} ($A_{pos}$) and \textbf{negative sentiment}($A_{neg}$).
    
\textbf{Compound sentiment} ($A_{compound}$). This combines the positive and negative sentiments to represent overall sentiment between -1 and 1.

\textbf{Valence} ($A_{valence}$) and \textbf{arousal} ($A_{arousal}$): Valence refers to the pleasantness of a stimulus and arousal refers to the intensity of emotion provoked by a stimulus \cite{warriner2013norms}. In \cite{aquino2007attention}, the authors specifically note that it is the arousal level of words, and not valence which is predictive of their effect on attention (measured via reaction time). Measuring early cortical responses and recall, \cite{kissler2007buzzwords} observed that words of greater valence were both more salient and memorable. To measure valence and arousal of a sentence, we use the averaged word rating, utilizing word ratings from the database introduced by \cite{warriner2013norms}.

\textbf{Concreteness} ($A_{concreteness}$): This is ``the degree to which the concept denoted by a word refers to a perceptible entity'' \cite{brysbaert2014concreteness}. As demonstrated by \cite{sadoski2000engaging}, concrete texts are better recalled than abstract ones and concreteness is a strong predictor of text comprehensibility, interest, and recall. To measure concreteness of a sentence, we use the averaged word rating, utilizing word ratings in the database introduced by \cite{brysbaert2014concreteness}.

\subsection{N-Gram Features}
\label{sec:models_ngram}

We consider character-level and word-level n-gram text representations, shown to perform well in related tasks \cite{potthast2016clickbait,chakraborty2016stop,lamprinidis2018predicting}. A passage of text is then represented by a vector of the counts of the individual n-grams it contains. We use a logistic regression classifier with these representations.

\subsection{SBERT Embeddings with a Progression of Neural Architectures}
\label{sec:models_embeddings}

\begin{figure}[h]
	\centering
	\includegraphics[width=1\textwidth]{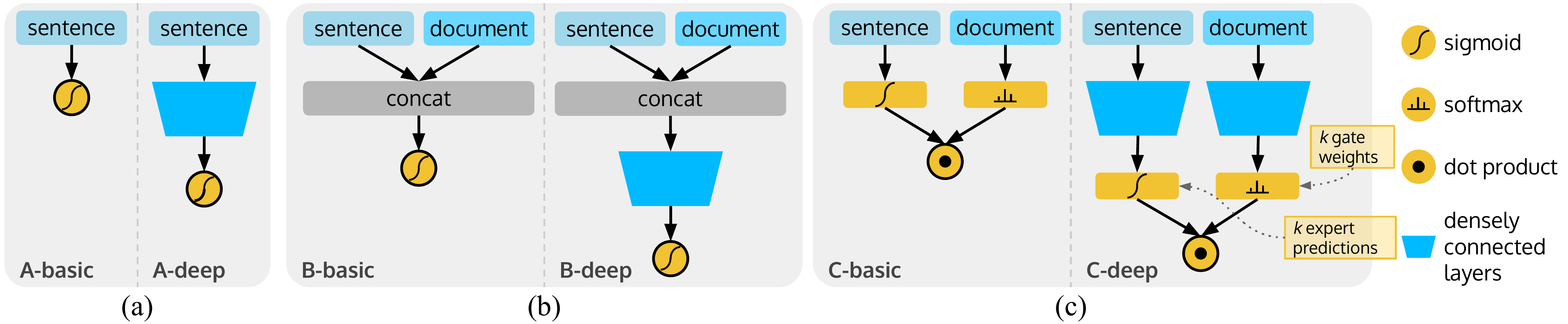}
	\caption{The progression of neural network architectures combined with SBERT sentence and document embeddings. Group A only uses sentence embeddings, while groups B and C also use document embeddings. In group C, they are combined in a mixture-of-experts fashion (the width of the sigmoid and softmax layers is equal to the \# experts). For each group, there is a basic version and deep version.}
	\label{fig:nn_progression}
\end{figure}


All other models described in this work use only the single sentence to predict PQ probability. To understand the importance of considering the entire article when choosing PQs, we consider three groups of neural architectures, as shown in Figure~\ref{fig:nn_progression}.

\textbf{Group A}. These neural networks only take the sentence embedding as input. In the \textbf{A-basic} model, there are no hidden layers. In \textbf{A-deep}, the embedding passes through a set of densely connected layers.

\textbf{Group B}. These models receive the sentence embedding and a whole-document embedding as input. This allows the models to account for document-dependent patterns. These embeddings are concatenated and connected to the output node (\textbf{B-basic}), or first pass through densely connected layers (\textbf{B-deep}).

\textbf{Group C}. These networks also receive sentence and document embeddings, but they are combined in a mixture-of-experts manner \cite{jacobs1991adaptive}. That is, multiple predictions are produced by a set of ``experts'' and a gating mechanism determines the weighting of these predictions for a given input. The motivation is that there may be many ``types'' of articles, each requiring paying attention to different properties when choosing a PQ. 
If each of $k$ experts generates a prediction, we can use the document embedding to determine the weighting over the predictions. In Figure~\ref{fig:nn_progression}c, $k$ corresponds to the width of the sigmoid and softmax layers, which are then combined with a dot product to produce the final prediction. In \textbf{C-deep}, the embeddings first pass through a set of densely connected layers (non-shared weights) as shown in the right of Figure~\ref{fig:nn_progression}c, while in \textbf{C-basic}, they do not.

To embed sentences and documents, we make use of a pre-trained Sentence-BERT (SBERT) model \cite{reimers2019sentence}. 
SBERT is a modification of BERT (Bidirectional Encoder Representations from Transformers) -- a language representation model which performs well on a wide variety of tasks \cite{devlin2018bert}. 
SBERT is designed to more efficiently produce semantically meaningful embeddings \cite{reimers2019sentence}. 
We computed document embeddings by averaging SBERT sentence embeddings.

\subsection{Cross-Task Models}
\label{sec:models_cross_task}

To test the similarity of PQ selection with related tasks
, we use the following models:
\textbf{Headline popularity}: We train a model to predict the popularity of a headline (using SBERT embeddings and linear regression) with the dataset introduced by \cite{moniz2018multi}. This dataset includes feedback metrics for about 100K news articles from various social media platforms. We apply this model to PQ selection by predicting the popularity of each sentence, scaling the predictions for each article to lie in $[0, 1]$ and interpreting these values as PQ probability.
\textbf{Clickbait identification}: We train a model to discriminate between clickbait and non-clickbait headlines (using SBERT embeddings and logistic regression) with the dataset introduced by \cite{chakraborty2016stop}. Clickbait probability is used as a proxy for PQ probability.
\textbf{Summarization}: Using a variety of extractive summarizers, we score each sentence in an article, scale the values to lie in $[0, 1]$, and interpret these values as PQ probability. No training is required for this model.
Appendix.~\ref{sec:implementation} contain implementation details of these models

\section{Experimental Setup}
\label{sec:setup}

To support the new task of automatic PQ selection, we both construct a new dataset and describe a suitable evaluation metric.

\subsection{Datatset Construction} 
\label{sec:dataset}

To conduct our experiments, we create a dataset using articles from several online news outlets: National Post, The Intercept, Ottawa Citizen, and Cosmopolitan. For each outlet, we identify those articles containing at least one pull quote. From these articles, we extract the \textit{body}, \textit{edited PQs}, and \textit{PQ source sentences}. The body contains the full list of sentences composing the body of the article. The edited PQs are the pulled texts as they appear after being augmented by the editor to appear as pull quotes\footnote{This can include replacing pronouns such as ``she'', ``they'', ``it'', with the more precise nouns or proper nouns, or shortening sentences by removing individual words or clauses, or even replacing words with ones of a similar meaning but different length in order to achieve a clean text rag.}. The PQ source sentences are the article sentences from which the edited PQs came. In this work, we aim to determine whether a given article sentence is a source sentence or not\footnote{A PQ source sentence could be only part of a multi-sentence PQ or contain the PQ inside it.}.

Dataset statistics are reoprted in Table \ref{tab:data_stats}. It contains $\sim$27K positive samples (PQ source sentences---which we simply call PQ sentences) and $\sim$680K negative samples (non-PQ sentences). The positive to negative ratio is 1:26 (taken into consideration when training our classifiers with balanced class weights). For all experiments, we use the same training/validation/test split of the articles (70/10/20).

\begin{table}[h]
\centering
\resizebox{\textwidth}{!}{
\begin{tabular}{@{}lrrrr|rrr|r@{}}
\toprule
 &
  \multicolumn{1}{c}{\textbf{nationalpost}} &
  \multicolumn{1}{c}{\textbf{theintercept}} &
  \multicolumn{1}{c}{\textbf{ottawacitizen}} &
  \multicolumn{1}{c|}{\textbf{cosmopolitan}} &
  \multicolumn{1}{c}{\textbf{train}} &
  \multicolumn{1}{c}{\textbf{val}} &
  \multicolumn{1}{c|}{\textbf{test}} &
  \multicolumn{1}{c}{\textbf{all}} \\ \midrule
\textbf{\# articles}          & 11112  & 1183   & 1066  & 1267  & 10239  & 1462  & 2927   & 14628  \\
\textbf{\# PQ}                & 16307  & 2671   & 1087  & 2360  & 15709  & 2235  & 4481   & 22425  \\
\textbf{\# PQ/article}        & 1.47   & 2.26   & 1.02  & 1.86  & 1.53   & 1.53  & 1.53   & 1.53   \\
\textbf{\# sentences/PQ}      & 1.16   & 1.23   & 1.32  & 1.24  & 1.19   & 1.18  & 1.19   & 1.19   \\
\textbf{\# sentences/article} & 40.49  & 97.94  & 38.35 & 79.03 & 48.47  & 47.8  & 48.06  & 48.32  \\
\textbf{\# pos samples}       & 18975  & 3274   & 1436  & 2906  & 18640  & 2625  & 5326   & 26591  \\
\textbf{\# neg samples}       & 430959 & 112588 & 39443 & 97230 & 477609 & 67258 & 135353 & 680220 \\ \bottomrule
\end{tabular}%
}
\caption{Statistics of our PQ dataset, composed of articles from four different news outlets. Only articles with at least one PQ are included in the dataset.}
\label{tab:data_stats}
\end{table}

\subsection{Evaluation} 
\label{sec:evaluation}

\textbf{What do we want to measure?} We want to evaluate a PQ selection model on its ability to determine which sentences are more likely to be chosen by an expert as PQ source sentences.

\textbf{Metric.} We will use the probability that a random PQ source sentence is scored by the model above a random non-source sentence from the same article (i.e. AUC). Let $a_{inclusions}$ be the binary vector indicating whether each sentence of article $a$ is truly a PQ source sentence, and let $\hat{a}_{inclusions}$ be the corresponding predicted probabilities. Our metric can then be computed with Equation~\ref{eqn:metric}, which computes the AUC averaged across articles.
\begin{equation}
\label{eqn:metric}
    AUC_{avg} = \frac{1}{\#articles} \sum_{a \in articles} AUC(a_{inclusions}, \hat{a}_{inclusions})
\end{equation}

\textbf{Why average across articles?} By averaging scores for each article instead of for all sentences at the same time, the evaluation method accounts for the observation that some articles may be more ``pull-quotable'' than others. If articles are instead combined when computing AUC, an average sentence from an interesting article can be ranked higher than the best sentence from a less interesting article.

\section{Experimental Results}
\label{sec:results}

We present our experimental results and analysis for the four groups of approaches: 
handcrafted features (Section~\ref{sec:results_hc}), 
n-gram features (Section~\ref{sec:results_ngram}), 
SBERT embeddings combined with a progression of neural architectures (Section~\ref{sec:results_embeddings}), 
and cross-task models (Section~\ref{sec:results_cross_task}). 
We also perform human evaluation of several models (Section~\ref{sec:human_eval}).
Appendix~\ref{sec:implementation} contains implementation details of our models, and Appendix~\ref{sec:app_examples} includes examples of PQ sentences selected by several models on various articles.

\begin{wrapfigure}[33]{r}{0.4\textwidth}
	\centering
	\subfigure[Performance of handcrafted features]{
    \includegraphics[width=1\textwidth]{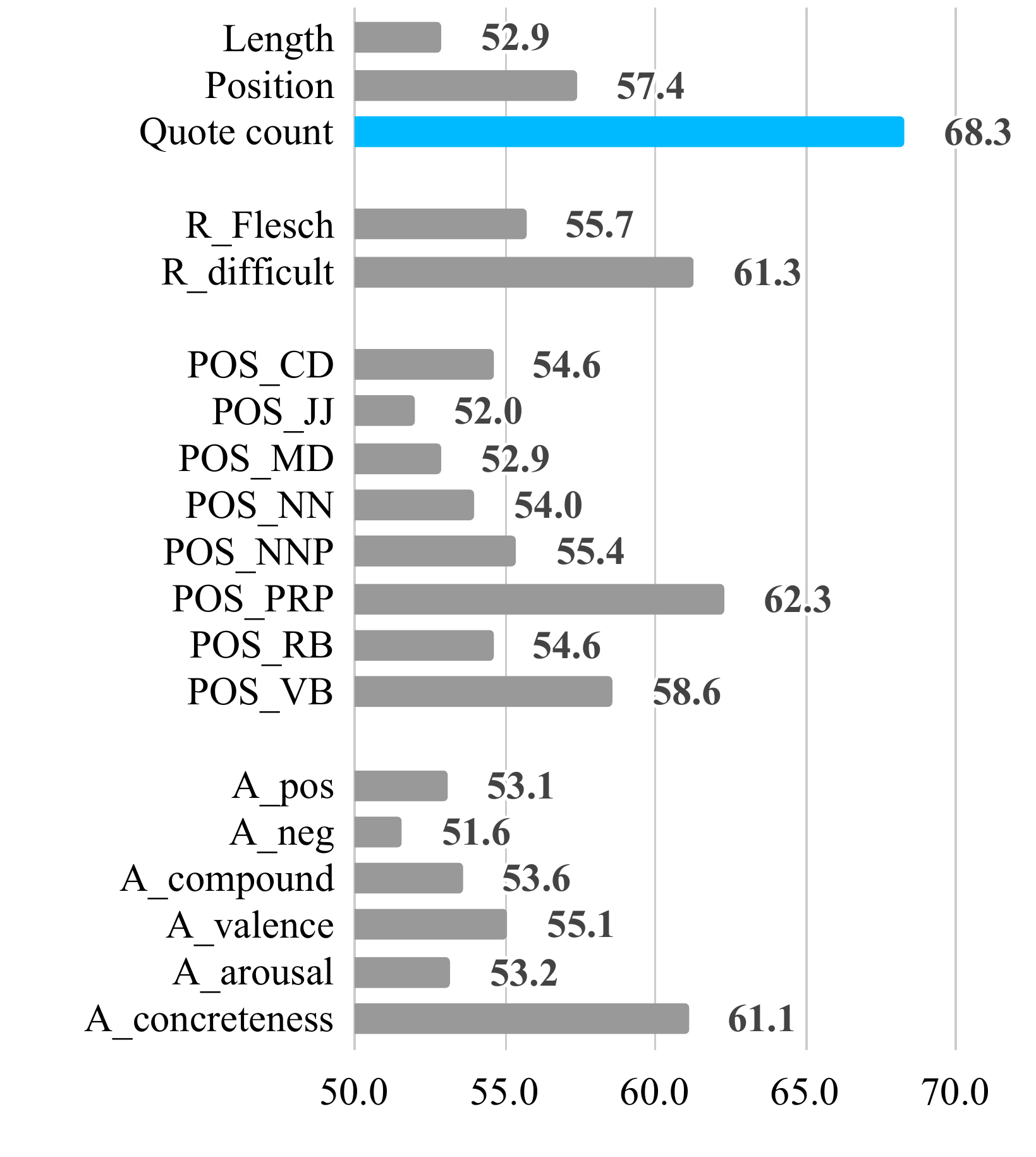}
    }
    
    \subfigure[Sentence position]{
    \includegraphics[width=.45\textwidth]{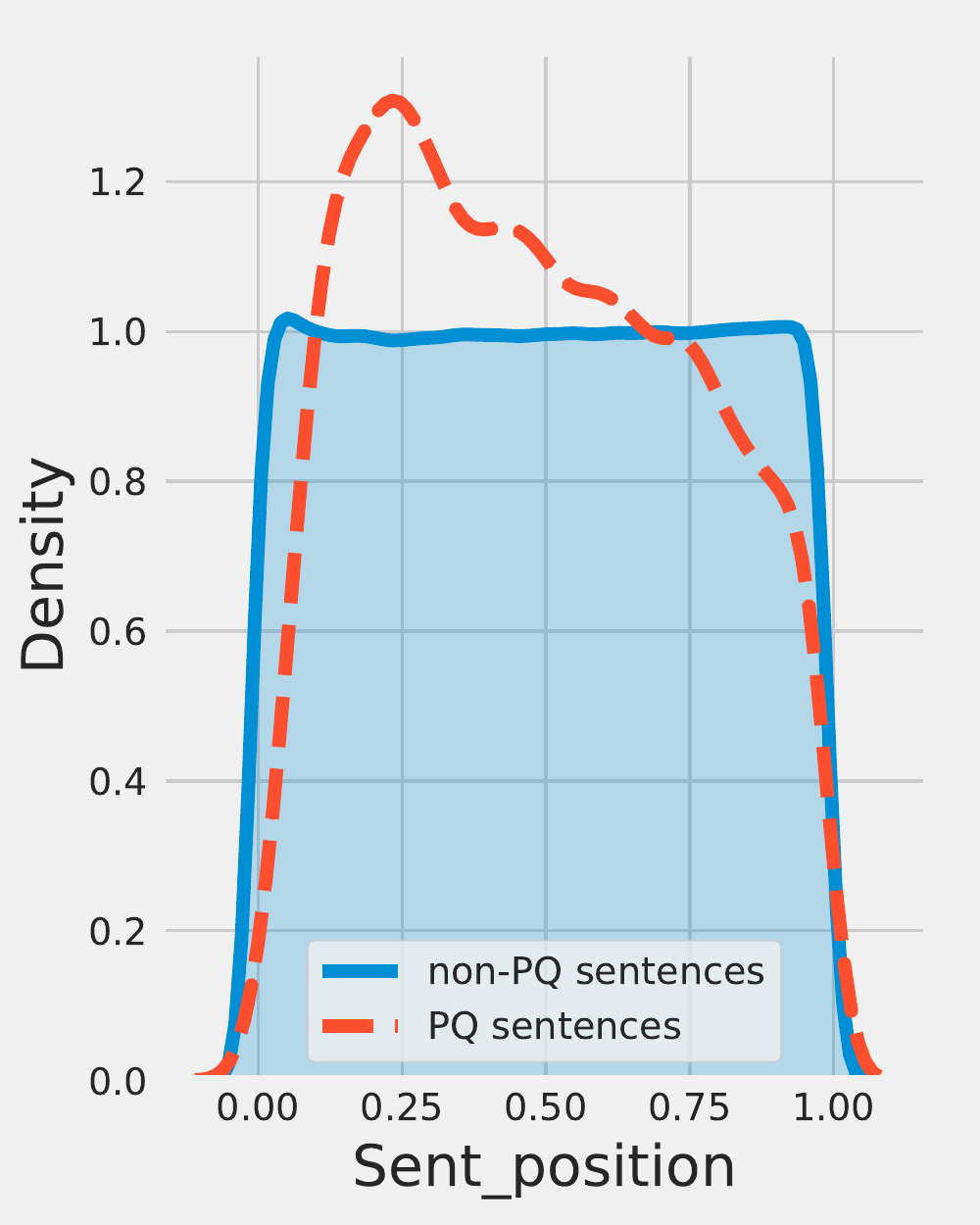}
    }
    \subfigure[Concreteness]{
    \includegraphics[width=.45\textwidth]{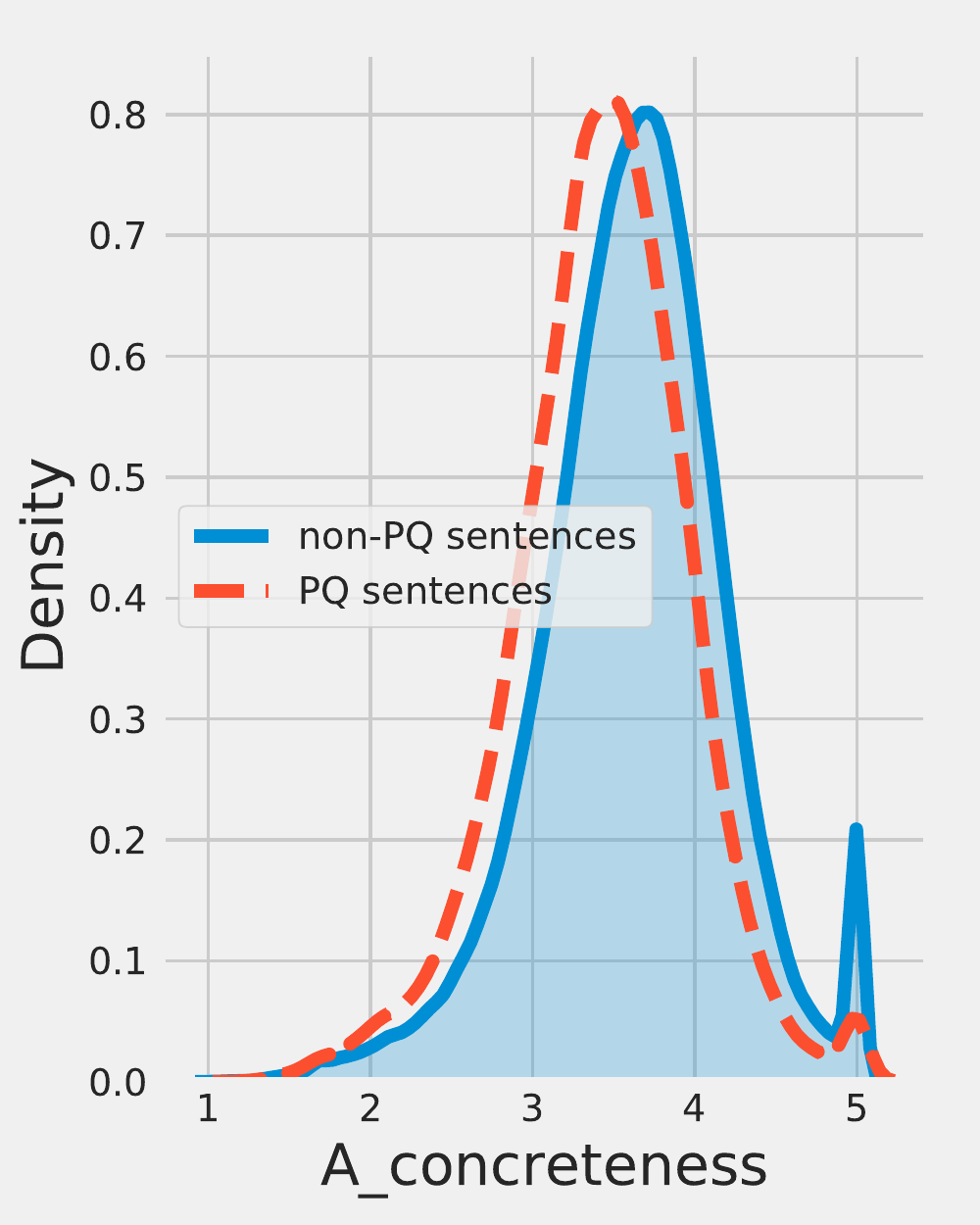}
    }
    \caption{The value distributions for two interesting handcrafted features for both non-PQ sentences (solid blue region) and PQ sentences (dashed orange lines).}
    \label{fig:hc_results}
\end{wrapfigure}

\subsection{Handcrafted Features}
\label{sec:results_hc}

The performance of each of our handcrafted features is provided in Figure~\ref{fig:hc_results}a. There are several interesting observations, including some that support and contradict hypotheses made in Section~\ref{sec:models_hc}:

\textbf{Sentence position}. Simply using the sentence position works better than random guessing. When we inspect the distribution of this feature value for PQ and non-PQ sentences in Figure~\ref{fig:hc_results}b, we see that PQ sentences are not uniformly distributed throughout articles, but rather tend to occur slightly more often around a quarter of the way through the article.
    
\textbf{Quotation mark count.}. The number of quotation marks is by far the best feature in this group, confirming that direct quotations make for good PQs. We find that a given non-PQ sentence is $\sim$3 times more likely not to contain quotation marks than a PQ sentence.
    
\textbf{Reading difficulty}. The fraction of difficult words is the third-best handcrafted feature, outperforming the Flesch metric. As suggested in Section~\ref{sec:models_hc_surface} we find that PQ sentences are indeed easier to read than non-PQ sentences.
    
\textbf{POS tags}. Of the POS tag densities, personal pronoun (PRP) and verb (VB) density are the most informative. Inspecting the feature distributions, we see that PQs tend to have slightly higher PRP density  as well as VB density -- suggesting that sentences about people doing things are good candidates for PQs.

\textbf{Affect features}. Affect features tended to perform poorly, contradicting our intuition that more exciting or emotional sentences would be chosen for PQs. However, concreteness is indeed an informative feature, with \textit{decreased} concreteness unexpectedly being better (see Figure~\ref{fig:hc_results}c). Given the memorability that comes with more concrete texts \cite{sadoski2000engaging}, this suggests that something else may be at work in order to explain the beneficial effects of PQs on learning outcomes \cite{wanta1994young,wanta1994information}.


\subsection{N-Gram Features}
\label{sec:results_ngram}

The results for our n-gram models are provided in Table~\ref{tab:results_ngram}. Impressively, almost all n-gram models performed better than any individual handcrafted feature, with the best model, character bi-grams, demonstrating an $AUC_{avg}$ of 75.4. When we inspect the learned logistic regression weights for the best variant of each model type (summarized in Figure~\ref{fig:top_grams}), we find a few interesting observations:

\begin{figure}
\begin{floatrow}
\capbtabbox{%
\resizebox{0.25\textwidth}{!}{%
    \begin{tabular}{@{}lrrr@{}}
    \toprule
    \textit{\textbf{Token}} & \textbf{n = 1} & \textbf{n = 2} & \textbf{n = 3} \\ \midrule
    \textbf{char}           & 70.7           & 75.4           & 74.2           \\
    \textbf{word}           & 73.9           & 72.3           & 65.6           \\ \bottomrule
    \end{tabular}%
}
}{%
  \caption{$AUC_{avg}$ scores of the n-gram models.}%
  \label{tab:results_ngram}
}
\ffigbox{%
  \includegraphics[width=0.4\textwidth]{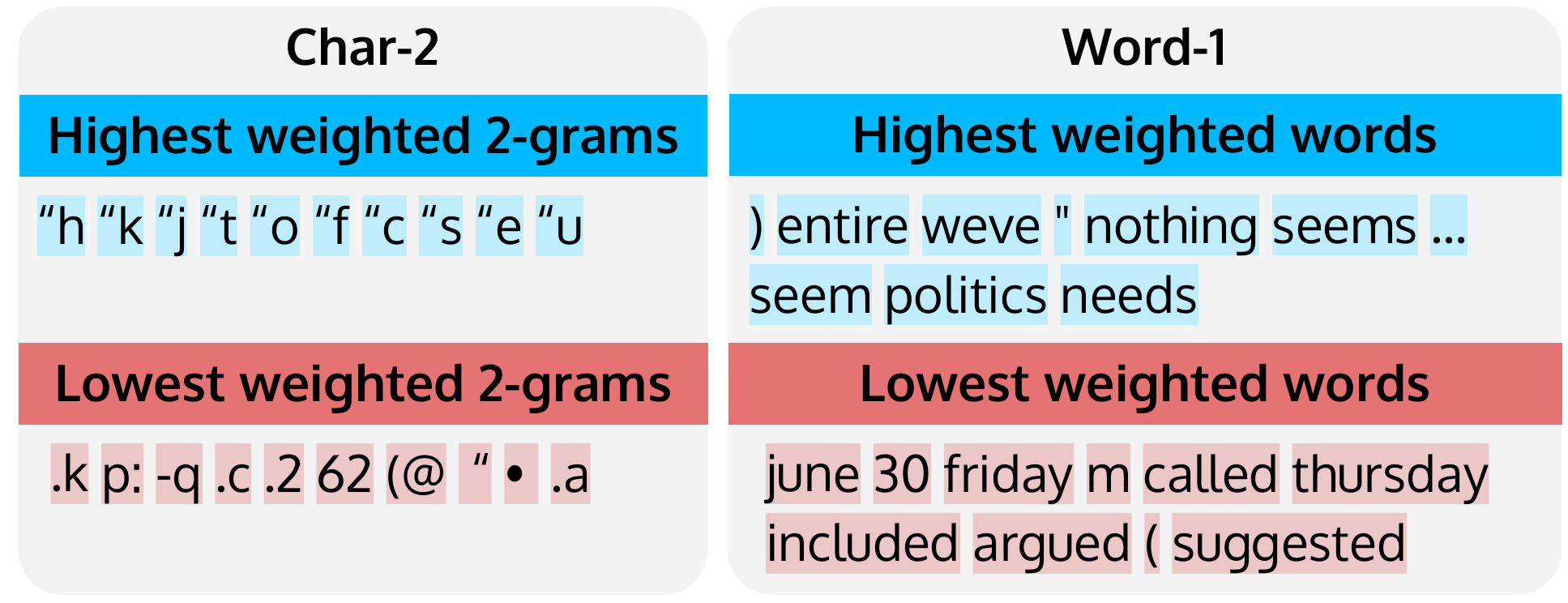}
}{%
  \caption{The ten highest and lowest weighted n-grams for the best character and word models.}%
  \label{fig:top_grams}
}
\end{floatrow}
\end{figure}

\textbf{Top character bi-grams}. The highest weighted character bi-grams exclusively aim to identify the beginnings of quotations, agreeing with the success of the quote count feature that the presence of a quote is highly informative. Curiously, the presence of a quotation being present but not starting the sentence is a strong negative indicator (i.e. `` ``''). 

\textbf{Bottom character bi-grams}. Among the lowest weighted character bi-grams are also indicators of numbers, URLs, and possibly twitter handles (i.e. ``@'').

\textbf{Words}. Although the highest weighted words are difficult to interpret together, among the lowest weighted words are those indicating past tense: ``called'', ``included'', ``argued'', ``suggested''. This suggests a promising approach for PQ selection includes identification of the tense of each sentence.
    


\subsection{SBERT Embeddings with a Progression of Neural Architectures}
\label{sec:results_embeddings}

\begin{wraptable}{r}{0.4\textwidth}

\centering
\resizebox{1\textwidth}{!}{%
\begin{tabular}{@{}lrcr@{}}
\toprule
\textbf{Model} & \multicolumn{1}{l}{$AUC_{avg}$} & \multicolumn{1}{l}{\textbf{Width}} & \multicolumn{1}{l}{\textbf{\# Params}} \\ \midrule
\textbf{A-basic}        & 76.7$\pm$0.15 & -                           & 7.7E+02 \\
\textbf{A-deep}         & 77.7$\pm$0.16 & \multicolumn{1}{r}{128, 64} & 1.1E+05 \\ \midrule
\textbf{B-basic}        & 77.1$\pm$0.24 & -                           & 1.5E+03 \\
\textbf{B-deep}         & 78.3$\pm$0.29 & \multicolumn{1}{r}{128, 64} & 2.1E+05 \\ \midrule
\textbf{C-basic ($k=16$)} & 77.7$\pm$0.51 & -                           & 2.5E+04 \\
\textbf{C-deep ($k=4$)}   & 78.7$\pm$0.07 & \multicolumn{1}{r}{32, 16}  & 5.0E+04 \\ \bottomrule
\end{tabular}%
}
\caption{Results on the neural architectures. Performance mean and std. dev. is calculated with five trials. $k$ refers to the \# experts, only applicable to C group models. Width values correspond to the width of the two additional fully connected layers (only applicable to the deep models).}
\label{tab:results_nn}
\end{wraptable}

The results of the neural architectures using SBERT embeddings is included in Table~\ref{tab:results_nn}. Overall, these results suggest that using document embeddings helps performance, especially with a mixture-of-experts architecture. This is seen by the general trend of improved performance from group A to B to C. Within each group, adding the fully connected layers (the ``deep'' models) helps.

\begin{figure}[ht]
	\centering
	\includegraphics[width=1\textwidth]{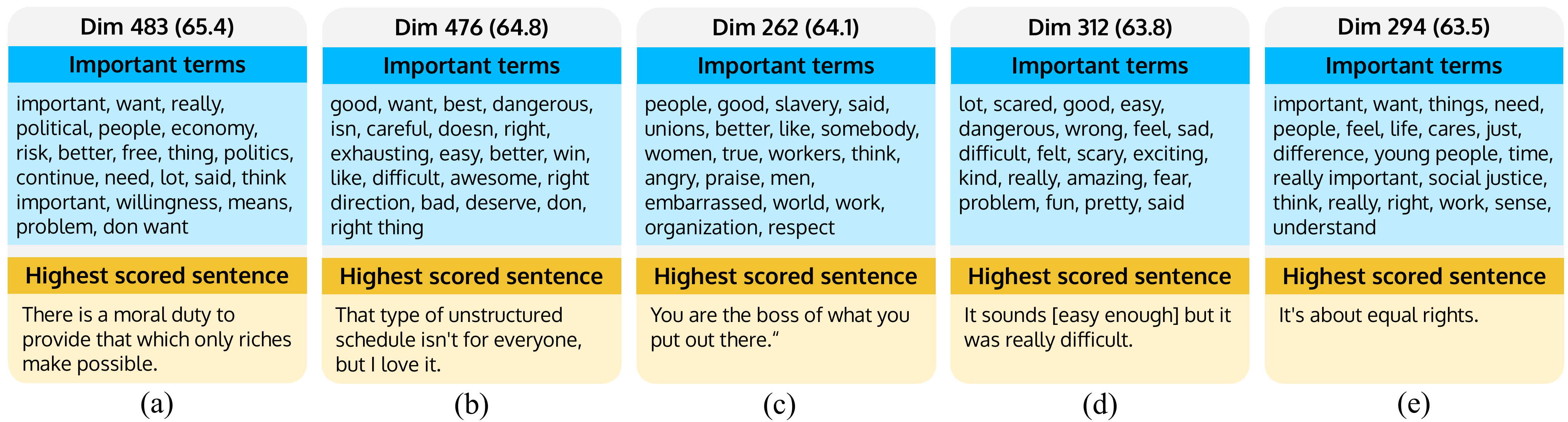}
	\caption{The top five best performing SBERT embedding dimensions, along with the terms associated with increasing PQ probability with respect to that dimension. For each dimension, we also include the sentence from the test articles which that dimension most strongly scores as being a PQ sentence. At the top of each box is the dimension index and the test $AUC_{avg}$.}
	\label{fig:dim_meanings}
\end{figure}

\textbf{Inspecting individual SBERT dimensions.} Given the performance of these embeddings, we are eager to understand what aspects of the text it picks up on. 
To do this, we first identify the most informative of the 768 dimensions for PQ selection by training a logistic regression model for each one.
For each single-feature model, we group sentences in the test set by PQ probability (high, medium, and low) and perform a TF-IDF analysis to identify key terms associated with \textit{increasing} PQ probability\footnote{Likewise, we could study terms associated with \textit{decreasing} PQ probability -- to deeper understand what \textit{bores} people.}. See Appendix~\ref{sec:app_tfidf} for more details. Results for the top five best performing dimensions are shown in Figure~\ref{fig:dim_meanings}. We find that each of these dimension is sensitive to the presence of a theme (or combination of themes) generally interesting and important to society. Our interpretations of them are: (a) politics and doing the right thing, (b) working hard on difficult/dangerous things, (c) discrimination, (d) strong emotions -- both positive and negative, and (e) social justice.

\subsection{Cross-Task Models}
\label{sec:results_cross_task}
%


\begin{wraptable}[11]{r}{0.25\textwidth}
\centering
\resizebox{1\textwidth}{!}{%
\begin{tabular}{@{}lr@{}}
\toprule
\textbf{Model}               & \multicolumn{1}{l}{$AUC_{avg}$} \\ \midrule
\textbf{headline popularity} & 56.9                                  \\ \midrule
\textbf{clickbait}           & 63.8                                  \\ \midrule
\textbf{LexRank}   & 51.9                                  \\
\textbf{SumBasic}  & 44.9                                  \\
\textbf{KLSum}        & 55.1                                  \\
\textbf{TextRank}  & 55.9                                  \\ \bottomrule
\end{tabular}%
}
\caption{Performance of the cross-task models.} 
\label{tab:results_cross_task}
\end{wraptable}

The results for the cross-task models of headline popularity prediction, clickbait identification, and summarization are shown in Table~\ref{tab:results_cross_task}.
Considered holistically, the results suggest that PQs are not designed to inform the reader about what they are reading (the shared purpose of headlines and summaries), so much as they are designed to motivate further engagement (the sole purpose of clickbait). However, the considerable performance gap between the clickbait model and PQ-specific models (such as character bi-grams and SBERT embeddings) suggest that this is only one aspect of choosing good pull quotes.

Another interesting observation is the variability in performance of summarizers at PQ selection. If we consider the summarization performance of these models as reported together in \cite{chen2016distraction}, we find that PQ selection performance is not strongly correlated with their summarization performance.

\subsection{Human Evaluation}
\label{sec:human_eval}

\begin{wraptable}{r}{0.45\textwidth}
\centering
\resizebox{1\textwidth}{!}{%
\begin{tabular}{@{}lrrr@{}}
\toprule
\textbf{Model} & \multicolumn{1}{l}{\textbf{Rating} $\uparrow$} & \multicolumn{1}{l}{\textbf{Rank} $\downarrow$} & \multicolumn{1}{l}{$\mathbf{1^{st}}$ \textbf{Place Pct.} $\uparrow$} \\ \midrule
\textbf{True PQ Source} & 2.75 & 3.04 & \textbf{28\%} \\ \midrule
\textbf{Char-2} & \textbf{2.86} & \textbf{2.74} & \textbf{28\%} \\
\textbf{C-deep} & 2.75 & 3.08 & 18\% \\ \midrule
\textbf{Headline pop.} & 2.57 & 3.66 & 8\% \\
\textbf{Clickbait} & 2.70 & 3.26 & 18\% \\
\textbf{TextRank} & 2.69 & 3.32 & 14\% \\ \bottomrule
\end{tabular}%
}
\caption{The results of human evaluation comparing models in terms of how interested the reader is in reading more of the article. The $\uparrow$ and $\downarrow$ indicate whether better values for a metric are respectively higher or lower.}
\label{tab:human_eval}
\end{wraptable}

As a final experiment, we conduct a qualitative evaluation to find out how well the PQs selected by various models (including the true PQ sources) compare. The results are summarized in Table~\ref{tab:human_eval}.
We randomly select 50 articles from the test set and ask nine volunteers to evaluate the candidate PQs extracted by six different models. They are asked to rate each of the 300 candidate PQs based on how interested it makes them in reading more of the article on a scale of 1 (not at all interested) to 5 (very interested). 
For each model we report the following metrics: (1) the \textbf{rating} averaged across all responses (with 5 being the best), (2) the average \textbf{rank} within an article (with 1 being the best), and (3) $\mathbf{1^{st}}$ \textbf{Place Pct.} -- how often the model produces the best PQ for an article (with 100\% being the best).

The results in Table~\ref{tab:human_eval} show that the two PQ-specific approaches (Char-2 and C-deep using the best hyperparameters from Section~\ref{sec:results_embeddings}) perform on par or slightly better than the true PQ sources. By generally out-performing the transfer models, this further supports our claim that the PQ selection task serves a unique purpose. When looking at how often each model scores $1^{st}$ place, which accentuates their performance differences, we can see that the headline and summarization models in particular perform poorly. Mirroring the results from Section~\ref{sec:results_cross_task}, among the cross-task models, the clickbait model seems to perform best.

\section{Conclusion}

In this work we proposed the novel task of automatic pull quote selection as a means to better understand how to engage readers.
To lay foundation for the task, we created a PQ dataset and described and benchmarked four groups of approaches: handcrafted features, n-grams, SBERT-based embeddings combined with a progression of neural architectures, and cross-task models.
By inspecting results, we encountered multiple curious findings to inspire further research on PQ selection and understanding reader engagement. 
%
%
%
%

There are many interesting avenues for future research with regard to pull quotes. In this work we assume that all true PQs in our dataset are of equal quality, however, it would be valuable to know the quality of individual PQs. It would also be interesting to study how to make a given phrase more PQ-worthy while maintaining the original meaning. 

\section*{Acknowledgements}
We acknowledge the support of the Natural Sciences and Engineering Research Council of Canada (NSERC) through the Discovery Grants Program. NSERC invests annually over \$1 billion in people, discovery and innovation.

\bibliographystyle{coling}
\bibliography{coling2020}

\begin{thebibliography}{}

\bibitem[\protect\citename{Aquino and Arnell}2007]{aquino2007attention}
Jennifer~M Aquino and Karen~M Arnell.
\newblock 2007.
\newblock Attention and the processing of emotional words: Dissociating effects
  of arousal.
\newblock {\em Psychonomic Bulletin \& Review}, 14(3):430--435.

\bibitem[\protect\citename{Bird \bgroup et al.\egroup }2009]{bird2009natural}
Steven Bird, Ewan Klein, and Edward Loper.
\newblock 2009.
\newblock {\em Natural Language Processing with Python: Analyzing Text with the
  Natural Language Toolkit}.
\newblock " O'Reilly Media, Inc.".

\bibitem[\protect\citename{Braddock}1974]{braddock1974frequency}
Richard Braddock.
\newblock 1974.
\newblock The frequency and placement of topic sentences in expository prose.
\newblock {\em Research in the Teaching of English}, 8(3):287--302.

\bibitem[\protect\citename{Brysbaert \bgroup et al.\egroup
  }2014]{brysbaert2014concreteness}
Marc Brysbaert, Amy~Beth Warriner, and Victor Kuperman.
\newblock 2014.
\newblock Concreteness ratings for 40 thousand generally known english word
  lemmas.
\newblock {\em Behavior Research Methods}, 46(3):904--911.

\bibitem[\protect\citename{Chakraborty \bgroup et al.\egroup
  }2016]{chakraborty2016stop}
Abhijnan Chakraborty, Bhargavi Paranjape, Sourya Kakarla, and Niloy Ganguly.
\newblock 2016.
\newblock Stop clickbait: Detecting and preventing clickbaits in online news
  media.
\newblock In {\em Advances in Social Networks Analysis and Mining (ASONAM),
  2016 IEEE/ACM International Conference on}, pages 9--16. IEEE.

\bibitem[\protect\citename{Chen \bgroup et al.\egroup
  }2016]{chen2016distraction}
Qian Chen, Xiaodan Zhu, Zhenhua Ling, Si~Wei, and Hui Jiang.
\newblock 2016.
\newblock Distraction-based neural networks for document summarization.
\newblock {\em arXiv preprint arXiv:1610.08462}.

\bibitem[\protect\citename{Chollet and others}2015]{chollet2015keras}
Fran\c{c}ois Chollet et~al.
\newblock 2015.
\newblock Keras.
\newblock \url{https://keras.io}.

\bibitem[\protect\citename{Devlin \bgroup et al.\egroup }2018]{devlin2018bert}
Jacob Devlin, Ming-Wei Chang, Kenton Lee, and Kristina Toutanova.
\newblock 2018.
\newblock Bert: Pre-training of deep bidirectional transformers for language
  understanding.
\newblock {\em arXiv preprint arXiv:1810.04805}.

\bibitem[\protect\citename{Erkan and Radev}2004]{erkan2004lexrank}
G{\"u}nes Erkan and Dragomir~R Radev.
\newblock 2004.
\newblock Lexrank: Graph-based lexical centrality as salience in text
  summarization.
\newblock {\em Journal of Artificial Intelligence Research}, 22:457--479.

\bibitem[\protect\citename{Flesch}1979]{flesch1979write}
Rudolf Flesch.
\newblock 1979.
\newblock {\em How to Write Plain English: A Book for Lawyers and Consumers}.
\newblock Harper \& Row New York, NY.

\bibitem[\protect\citename{French}2018]{french2018indesign}
Nigel French.
\newblock 2018.
\newblock {\em InDesign Type: Professional Typography with Adobe InDesign}.
\newblock Adobe Press.

\bibitem[\protect\citename{Haghighi and
  Vanderwende}2009]{haghighi2009exploring}
Aria Haghighi and Lucy Vanderwende.
\newblock 2009.
\newblock Exploring content models for multi-document summarization.
\newblock In {\em Proceedings of Human Language Technologies: The 2009 Annual
  Conference of the North American Chapter of the Association for Computational
  Linguistics}, pages 362--370.

\bibitem[\protect\citename{Hasan and Ng}2014]{hasan2014automatic}
Kazi~Saidul Hasan and Vincent Ng.
\newblock 2014.
\newblock Automatic keyphrase extraction: A survey of the state of the art.
\newblock In {\em Proceedings of the 52nd Annual Meeting of the Association for
  Computational Linguistics (Volume 1: Long Papers)}, pages 1262--1273,
  Baltimore, Maryland, June. Association for Computational Linguistics.

\bibitem[\protect\citename{Hastie \bgroup et al.\egroup }2009]{hastie2009multi}
Trevor Hastie, Saharon Rosset, Ji~Zhu, and Hui Zou.
\newblock 2009.
\newblock Multi-class adaboost.
\newblock {\em Statistics and its Interface}, 2(3):349--360.

\bibitem[\protect\citename{Hinton \bgroup et al.\egroup
  }2012]{hinton2012improving}
Geoffrey~E Hinton, Nitish Srivastava, Alex Krizhevsky, Ilya Sutskever, and
  Ruslan~R Salakhutdinov.
\newblock 2012.
\newblock Improving neural networks by preventing co-adaptation of feature
  detectors.
\newblock {\em arXiv preprint arXiv:1207.0580}.

\bibitem[\protect\citename{Holmes}2015]{holmes2015subediting}
Tim Holmes.
\newblock 2015.
\newblock {\em Subediting and Production for Journalists: Print, Digital \&
  Social}.
\newblock Routledge.

\bibitem[\protect\citename{Hutto and Gilbert}2014]{hutto2014vader}
Clayton~J Hutto and Eric Gilbert.
\newblock 2014.
\newblock Vader: A parsimonious rule-based model for sentiment analysis of
  social media text.
\newblock In {\em Eighth International AAAI Conference on Weblogs and Social
  Media}.

\bibitem[\protect\citename{Jacobs \bgroup et al.\egroup
  }1991]{jacobs1991adaptive}
Robert~A Jacobs, Michael~I Jordan, Steven~J Nowlan, and Geoffrey~E Hinton.
\newblock 1991.
\newblock Adaptive mixtures of local experts.
\newblock {\em Neural Computation}, 3(1):79--87.

\bibitem[\protect\citename{Kingma and Ba}2014]{kingma2014adam}
Diederik~P Kingma and Jimmy Ba.
\newblock 2014.
\newblock Adam: A method for stochastic optimization.
\newblock {\em arXiv preprint arXiv:1412.6980}.

\bibitem[\protect\citename{Kissler \bgroup et al.\egroup
  }2007]{kissler2007buzzwords}
Johanna Kissler, Cornelia Herbert, Peter Peyk, and Markus Junghofer.
\newblock 2007.
\newblock Buzzwords: Early cortical responses to emotional eords during
  reading.
\newblock {\em Psychological Science}, 18(6):475--480.

\bibitem[\protect\citename{Klambauer \bgroup et al.\egroup
  }2017]{klambauer2017self}
G{\"u}nter Klambauer, Thomas Unterthiner, Andreas Mayr, and Sepp Hochreiter.
\newblock 2017.
\newblock Self-normalizing neural networks.
\newblock In {\em Advances in Neural Information Processing Systems}, pages
  971--980.

\bibitem[\protect\citename{Lamprinidis \bgroup et al.\egroup
  }2018]{lamprinidis2018predicting}
Sotiris Lamprinidis, Daniel Hardt, and Dirk Hovy.
\newblock 2018.
\newblock Predicting news headline popularity with syntactic and semantic
  knowledge using multi-task learning.
\newblock In {\em Proceedings of the 2018 Conference on Empirical Methods in
  Natural Language Processing}, pages 659--664.

\bibitem[\protect\citename{Liu \bgroup et al.\egroup }2009]{liu2009clustering}
Zhiyuan Liu, Peng Li, Yabin Zheng, and Maosong Sun.
\newblock 2009.
\newblock Clustering to find exemplar terms for keyphrase extraction.
\newblock In {\em Proceedings of the 2009 Conference on Empirical Methods in
  Natural Language Processing: Volume 1-Volume 1}, pages 257--266. Association
  for Computational Linguistics.

\bibitem[\protect\citename{Luhn}1958]{luhn1958automatic}
Hans~Peter Luhn.
\newblock 1958.
\newblock The automatic creation of literature abstracts.
\newblock {\em IBM Journal of Research and Development}, 2(2):159--165.

\bibitem[\protect\citename{Medelyan \bgroup et al.\egroup
  }2009]{medelyan2009human}
Olena Medelyan, Eibe Frank, and Ian~H Witten.
\newblock 2009.
\newblock Human-competitive tagging using automatic keyphrase extraction.
\newblock In {\em Proceedings of the 2009 Conference on Empirical Methods in
  Natural Language Processing: Volume 3-Volume 3}, pages 1318--1327.
  Association for Computational Linguistics.

\bibitem[\protect\citename{Mihalcea and Tarau}2004]{mihalcea2004textrank}
Rada Mihalcea and Paul Tarau.
\newblock 2004.
\newblock Textrank: Bringing order into text.
\newblock In {\em Proceedings of the 2004 Conference on Empirical Methods in
  Natural Language Processing}, pages 404--411.

\bibitem[\protect\citename{Moniz and Torgo}2018]{moniz2018multi}
Nuno Moniz and Lu{\'\i}s Torgo.
\newblock 2018.
\newblock Multi-source social feedback of online news feeds.
\newblock {\em arXiv preprint arXiv:1801.07055}.

\bibitem[\protect\citename{Nallapati \bgroup et al.\egroup
  }2016a]{nallapati2016abstractive}
Ramesh Nallapati, Bowen Zhou, Caglar Gulcehre, Bing Xiang, et~al.
\newblock 2016a.
\newblock Abstractive text summarization using sequence-to-sequence rnns and
  beyond.
\newblock {\em arXiv preprint arXiv:1602.06023}.

\bibitem[\protect\citename{Nallapati \bgroup et al.\egroup
  }2016b]{nallapati2016classify}
Ramesh Nallapati, Bowen Zhou, and Mingbo Ma.
\newblock 2016b.
\newblock Classify or select: Neural architectures for extractive document
  summarization.
\newblock {\em arXiv preprint arXiv:1611.04244}.

\bibitem[\protect\citename{Nallapati \bgroup et al.\egroup
  }2017]{nallapati2017summarunner}
Ramesh Nallapati, Feifei Zhai, and Bowen Zhou.
\newblock 2017.
\newblock Summarunner: A recurrent neural network based sequence model for
  extractive summarization of documents.
\newblock In {\em Thirty-First AAAI Conference on Artificial Intelligence}.

\bibitem[\protect\citename{Nenkova and McKeown}2012]{nenkova2012survey}
Ani Nenkova and Kathleen McKeown.
\newblock 2012.
\newblock A survey of text summarization techniques.
\newblock In {\em Mining Text Data}, pages 43--76. Springer.

\bibitem[\protect\citename{Nenkova and Vanderwende}2005]{nenkova2005impact}
Ani Nenkova and Lucy Vanderwende.
\newblock 2005.
\newblock The impact of frequency on summarization.
\newblock {\em Microsoft Research, Redmond, Washington, Tech. Rep.
  MSR-TR-2005}, 101.

\bibitem[\protect\citename{Nenkova \bgroup et al.\egroup
  }2011]{nenkova2011automatic}
Ani Nenkova, Kathleen McKeown, et~al.
\newblock 2011.
\newblock Automatic summarization.
\newblock {\em Foundations and Trends{\textregistered} in Information
  Retrieval}, 5(2--3):103--233.

\bibitem[\protect\citename{Pedregosa \bgroup et al.\egroup
  }2011]{pedregosa2011scikit}
Fabian Pedregosa, Ga{\"e}l Varoquaux, Alexandre Gramfort, Vincent Michel,
  Bertrand Thirion, Olivier Grisel, Mathieu Blondel, Peter Prettenhofer, Ron
  Weiss, Vincent Dubourg, et~al.
\newblock 2011.
\newblock Scikit-learn: Machine learning in python.
\newblock {\em Journal of Machine Learning Research}, 12(Oct):2825--2830.

\bibitem[\protect\citename{Piotrkowicz \bgroup et al.\egroup
  }2017]{piotrkowicz2017headlines}
Alicja Piotrkowicz, Vania Dimitrova, Jahna Otterbacher, and Katja Markert.
\newblock 2017.
\newblock Headlines matter: Using headlines to predict the popularity of news
  articles on twitter and facebook.
\newblock In {\em Eleventh International AAAI Conference on Web and Social
  Media}.

\bibitem[\protect\citename{Potthast \bgroup et al.\egroup
  }2016]{potthast2016clickbait}
Martin Potthast, Sebastian K{\"o}psel, Benno Stein, and Matthias Hagen.
\newblock 2016.
\newblock Clickbait detection.
\newblock In {\em European Conference on Information Retrieval}, pages
  810--817. Springer.

\bibitem[\protect\citename{Reimers and Gurevych}2019]{reimers2019sentence}
Nils Reimers and Iryna Gurevych.
\newblock 2019.
\newblock Sentence-bert: Sentence embeddings using siamese bert-networks.
\newblock {\em arXiv preprint arXiv:1908.10084}.

\bibitem[\protect\citename{Romary}2010]{romary2010automatic}
Patrice Lopez~Laurent Romary.
\newblock 2010.
\newblock Automatic key term extraction from scientific articles in grobid.
\newblock In {\em SemEval 2010 Workshop}, page~4.

\bibitem[\protect\citename{Sadoski \bgroup et al.\egroup
  }2000]{sadoski2000engaging}
Mark Sadoski, Ernest~T Goetz, and Maximo Rodriguez.
\newblock 2000.
\newblock Engaging texts: Effects of concreteness on comprehensibility,
  interest, and recall in four text types.
\newblock {\em Journal of Educational Psychology}, 92(1):85.

\bibitem[\protect\citename{Schupp \bgroup et al.\egroup
  }2007]{schupp2007selective}
Harald~T Schupp, Jessica Stockburger, Maurizio Codispoti, Markus Jungh{\"o}fer,
  Almut~I Weike, and Alfons~O Hamm.
\newblock 2007.
\newblock Selective visual attention to emotion.
\newblock {\em Journal of Neuroscience}, 27(5):1082--1089.

\bibitem[\protect\citename{Stovall}1997]{stovall1997infographics}
James~Glen Stovall.
\newblock 1997.
\newblock {\em Infographics: A Journalist's Guide}.
\newblock Allyn \& Bacon.

\bibitem[\protect\citename{Tomokiyo and Hurst}2003]{tomokiyo2003language}
Takashi Tomokiyo and Matthew Hurst.
\newblock 2003.
\newblock A language model approach to keyphrase extraction.
\newblock In {\em Proceedings of the ACL 2003 Workshop on Multiword
  Expressions: Analysis, Acquisition and Treatment}, pages 33--40.

\bibitem[\protect\citename{Turney}1999]{turney1999learning}
Peter Turney.
\newblock 1999.
\newblock Learning to extract key phrases from text, nrc technical report
  erb{\textperiodcentered} 1057.
\newblock Technical report, Canada: National Research Council.

\bibitem[\protect\citename{Venneti and Alam}2018]{venneti2018curiosity}
Lasya Venneti and Aniket Alam.
\newblock 2018.
\newblock How curiosity can be modeled for a clickbait detector.
\newblock {\em arXiv preprint arXiv:1806.04212}.

\bibitem[\protect\citename{Wanta and Gao}1994]{wanta1994young}
Wayne Wanta and Dandan Gao.
\newblock 1994.
\newblock Young readers and the newspaper: Information recall and perceived
  enjoyment, readability, and attractiveness.
\newblock {\em Journalism Quarterly}, 71(4):926--936.

\bibitem[\protect\citename{Wanta and Remy}1994]{wanta1994information}
Wayne Wanta and Jay Remy.
\newblock 1994.
\newblock Information recall of four newspaper elements among young readers.

\bibitem[\protect\citename{Warriner \bgroup et al.\egroup
  }2013]{warriner2013norms}
Amy~Beth Warriner, Victor Kuperman, and Marc Brysbaert.
\newblock 2013.
\newblock Norms of valence, arousal, and dominance for 13,915 english lemmas.
\newblock {\em Behavior Research Methods}, 45(4):1191--1207.

\bibitem[\protect\citename{Zhang \bgroup et al.\egroup }2019]{zhang2019pegasus}
Jingqing Zhang, Yao Zhao, Mohammad Saleh, and Peter~J Liu.
\newblock 2019.
\newblock Pegasus: Pre-training with extracted gap-sentences for abstractive
  summarization.
\newblock {\em arXiv preprint arXiv:1912.08777}.

\end{thebibliography}

\begin{appendices}
\counterwithin{table}{section}
\section{Implementation Details}
\label{sec:implementation}

Here we outline the various tools, datasets, and other implementation details related to our experiments:
\begin{itemize}
    \item To perform part-of-speech tagging for feature extraction, we use the NLTK 3.4.5 perceptron tagger \cite{bird2009natural}.
    
    \item To compute sentiment, the VADER Sentiment Analysis tool is used \cite{hutto2014vader}, accessed through the NLTK library.
    
   
    \item Implementations of the $R_{Flesch}$ readability metric is provided by the Textstat 0.6.0 Python package\footnote{Available online here: \url{https://github.com/shivam5992/textstat}}. The corpus of easy words for $R_{difficult}$ is also made available by this package.
    
    \item Valence, arousal word ratings are obtained from the dataset described in \cite{warriner2013norms}\footnote{Available online at \url{http://crr.ugent.be/archives/1003}.}. When computing average valence and arousal for a sentence, stop words are removed and when a word rating cannot be found, a value of 5 is used for valence and 4 for arousal (the mean word ratings).
    
    \item Concreteness word ratings are obtained from the dataset described in \cite{brysbaert2014concreteness} \footnote{Available online at \url{http://crr.ugent.be/archives/1330}.}. The concreteness score of a sentence is computed similar to valence and arousal, with a mean word rating of 5 used when no value for a word is available.
    
    \item For the n-gram models, a vocabulary size of 1000 was used for all models, and lower-casing was applied for the character and word models.
    
    \item The SBERT \cite{reimers2019sentence} implementation and pre-trained models are used for text embedding\footnote{Can be found online at \url{https://github.com/UKPLab/sentence-transformers}. We use the \texttt{bert-base-nli-mean-tokens} pre-trained model.}. 
    
    \item All neural networks using the SBERT embeddings were implemented with the Keras library \cite{chollet2015keras} with the Adam optimizer \cite{kingma2014adam} (with default Keras settings) and binary cross-entropy loss. Early stopping is done after validation loss stops decreasing for 4 epochs -- with a maximum of 100 epochs. In the deep version of the models, we include two additional densely connected layers as shown in Figure~\ref{fig:nn_progression}, with the second additional layer having half the width of the initial one. We use selu activations \cite{klambauer2017self} for the additional layers and a dropout rate of 0.5 for only the first additional densely connected layer \cite{hinton2012improving}. The hyperparameters requiring tuning for each model and the range of values tested (grid search) is provided in Table~\ref{tab:nn_hyperparams}.
    

    \item The clickbait identification dataset introduced by \cite{chakraborty2016stop} is used, which contains 16,000 clickbait samples and 16,000 non-clickbait headlines\footnote{Available online at \url{https://github.com/bhargaviparanjape/clickbait/tree/master/dataset}.}.
    
    \item The headline popularity dataset introduced by \cite{moniz2018multi} is used, which includes feedback metrics for about 100,000 news articles from various social media platforms\footnote{Available online at \url{https://archive.ics.uci.edu/ml/machine-learning-databases/00432/Data/}.}. For pre-processing, we remove those article where no popularity feedback data is available, and compute popularity by averaging percentiles across platforms. For example, if an article is in the $80^{th}$ popularity percentile on Facebook and in the $90^{th}$ percentile on LinkedIn, then it is given a popularity score of 0.85.
    
    \item We use the following summarizers: TextRank \cite{mihalcea2004textrank}, SumBasic \cite{nenkova2005impact}, LexRank \cite{erkan2004lexrank}, and KLSum \cite{haghighi2009exploring}\footnote{Implementations provided by Sumy library, available at \url{https://pypi.python.org/pypi/sumy}.}.
    

    \item We used the Scikit-learn \cite{pedregosa2011scikit} implementations of AdaBoost, decision trees, and logistic regression. To accommodate the imbalanced training data, balanced class weighting was used for the decision trees in Adaboost and logistic regression. For AdaBoost, we use 100 estimators with the default learning rate of 1.0. For logistic regression we use the default settings of L2 penalty with $C=1.0$.

\end{itemize}

\begin{table}[ht]
\centering
\resizebox{0.4\textwidth}{!}{%
\begin{tabular}{@{}lcc@{}}
\toprule
\textbf{model}   & \textbf{Initial width}          & \textbf{\# Experts} \\ \midrule
\textbf{A-basic} & -                               & -                   \\
\textbf{A-deep}  & {[}16, 32, 64, 128, 256, 512{]} & -                   \\ \midrule
\textbf{B-basic} & -                               & -                   \\
\textbf{B-deep}  & {[}16, 32, 64, 128, 256, 512{]} & -                   \\ \midrule
\textbf{C-basic} & -                               & {[}2, 4, 8, 16{]}   \\
\textbf{C-deep}  & {[}16, 32, 64, 128, 256, 512{]} & {[}2, 4, 8, 16{]}   \\ \bottomrule
\end{tabular}%
}
\caption{Hyperparameter values used in grid search for the different SBERT neural networks. The models with the best performance on the validation set averaged across 5 trials are reported in Table~\ref{tab:results_nn}.}
\label{tab:nn_hyperparams}
\end{table}

\section{TF-IDF Analysis of SBERT Embedding Dimensions}
\label{sec:app_tfidf}

In order to uncover the key terms associated with increased PQ probability for a given SBERT embedding dimension, the following steps were performed:
\begin{enumerate}
    \item Train a logistic regression model using that single feature. Make a note of whether the coefficient is positive (i.e. increasing the feature value increase PQ probability) or negative (i.e. decreasing feature value increases PQ probability).
    \item Take all test sentences and split them into three groups: (1) those where the feature value is in the top $k$, (2) those where the feature value is in the middle $2k$, and (3) those where the feature value is in the bottom $k$. We use $k = 2000$.
    \item Join together the sentences within each of the three groups so that we have three ``documents'' and apply TF-IDF on this set of documents. We use the Scikit-learn \cite{pedregosa2011scikit} implementation, with an n-gram range of 1-3 words and use the English stopword list with \texttt{sublinear\_tf = True}. All other settings are at the default values.
    \item If the coefficient from step 1 is positive, use the highest ranked terms for group 1. If the coefficient is negative, use the highest ranked terms for group 3.
\end{enumerate}

\section{Model-Chosen Pull Quote Examples}
\label{sec:app_examples}

\begin{table}[h]
\centering
\resizebox{\textwidth}{!}{%
\begin{tabular}{@{}ll@{}}
\toprule
\textbf{Model}               & \textbf{Highest rated sentence(s)}                                                                                          \\ \midrule
\textbf{True PQ Source} & \cellcolor[HTML]{9AFF99}“To date, the fishing industry in British Columbia has not raised the carbon tax as an area of specific concern,” it \\
\textbf{}                    & \cellcolor[HTML]{9AFF99}\quad says.                                                                                           \\ \midrule
\textbf{Quote\_count}        & OTTAWA  -  The federal government’s carbon tax could take a toll on Canada’s fishing industry, causing its                  \\
\textbf{}                    & \quad competitiveness to “degrade relative to other nations,” according to an analysis from the fisheries department.         \\
\textbf{Sent\_position} & In the aquaculture and seafood processing industries, in contrast, fuel makes up just 1.6 per cent and 0.8 per cent of                       \\
\textbf{}                    & \quad total costs, respectively.                                                                                              \\
\textbf{R\_difficult}        & That would result in a difference in the GDP of about \$2 billion in 2022, or 0.1 per cent.                                 \\
\textbf{POS\_PRP}            & “To date, the fishing industry in British Columbia has not raised the carbon tax as an area of specific concern,” it        \\
\textbf{}                    & \quad says.                                                                                                                   \\
\textbf{POS\_VB}             & “The relatively rapid introduction of measures to reduce GHG emissions would allow little time for industry and             \\
\textbf{}                    & \quad consumers to adjust their behaviour, creating a substantial risk of economic disruption and uncertainty.”               \\
\textbf{A\_concreteness}     & “This could have a negative impact on the competitiveness of Canada’s fishing industry.”                                    \\ \midrule
\textbf{Char-2}              & “However, Canada’s competitiveness may degrade relative to other nations that have not yet announced plans, or are          \\
\textbf{}                    & \quad proceeding more slowly towards measures to reduce GHG emissions,” the memo says.                                        \\
\textbf{Word-1}              & The memo concludes that short-term impacts are expected to be “low to moderate,” and the department will “continue to       \\
\textbf{}                    & \quad monitor developments.”                                                                                                  \\ \midrule
\textbf{C-deep}              & “To date, the fishing industry in British Columbia has not raised the carbon tax as an area of specific concern,” it        \\
\textbf{}                    & \quad says.                                                                                                                   \\ \midrule
\textbf{Headline popularity} & The four largest provinces  -  Quebec, Ontario, Alberta and B.C.                                                            \\
\textbf{Clickbait}           & Ottawa has said all jurisdictions that don’t have their own carbon pricing plans in place this year will have the           \\
\textbf{}                    & \quad federal carbon tax imposed on them in January 2019, starting at $20 per tonne and increasing to $50 per tonne in 2022.  \\
\textbf{TextRank}            & The analysis was completed in December 2016, shortly after most provinces and territories had signed Ottawa’s pan-          \\
\textbf{}                    & \quad Canadian climate change framework, committing them to a range of measures, including carbon pricing, to reduce Canada’s \\
\textbf{}                    & \quad 2030 emissions to 30 per cent below 2005 levels.                                                                        \\ \bottomrule
\end{tabular}%
}
\caption{Article source: \url{https://nationalpost.com/news/politics/federal-carbon-tax-could-degrade-canadian-fishing-industrys-competitiveness-says-memo}.}
\label{tab:pq_samples_1}
\end{table}

\begin{table}[h]
\centering
\resizebox{\textwidth}{!}{%
\begin{tabular}{@{}ll@{}}
\toprule
\textbf{Model}               & \textbf{Highest rated sentence(s)}                                                                                     \\ \midrule
\textbf{True PQ Source}      & \cellcolor[HTML]{9AFF99}I think so many people voted for me because I think they’re just proud of me as well.          \\ \midrule
\textbf{Quote\_count}        & The school year is finally coming to an end and that means it's prom season, woo season!                               \\
\textbf{Sent\_position}      & I texted my friends like, “Oh my god I’m freaking out.                                                                 \\
\textbf{R\_difficult}        & I'm only at the school for an hour and a half every other day so I had no idea that we were even voting.               \\
\textbf{POS\_PRP}            & I think so many people voted for me because I think they’re just proud of me as well.                                  \\
\textbf{POS\_VB}             & -  and some people would send me them, but I just choose not to read them.                                             \\
\textbf{A\_concreteness}     & I didn't hear about anything.                                                                                          \\ \midrule
\textbf{Char-2}              & Something that I just want everyone to take away from this is you can be you as long as you're not hurting anyone else \\
\textbf{}                    & \quad and as long as you're not breaking any rules.                                                                      \\
\textbf{Word-1}              & Something that I just want everyone to take away from this is you can be you as long as you're not hurting anyone else \\
\textbf{}                    & \quad and as long as you're not breaking any rules.                                                                      \\ \midrule
\textbf{C-deep}              & I don't think there's any day where I haven't worn a full face of makeup to school, and I always dress up.             \\ \midrule
\textbf{Headline popularity} & I think so many people voted for me because I think they’re just proud of me as well.                                  \\
\textbf{Clickbait}           & I texted my friends like, “Oh my god I’m freaking out.                                                                 \\
\textbf{TextRank} & In an interview with Cosmopolitan.com, he talked about putting together his look, why he didn't see his crowning coming, \\
\textbf{}                    & \quad and what he'd like to tell the haters.                                                                             \\ \bottomrule
\end{tabular}%
}
\caption{Article source: \url{https://www.cosmopolitan.com/lifestyle/a20107039/south-carolina-prom-king-adam-bell-interview/}}
\label{tab:pq_samples_2}
\end{table}

\begin{table}[h]
\centering
\resizebox{\textwidth}{!}{%
\begin{tabular}{@{}ll@{}}
\toprule
\textbf{Model}              & \textbf{Highest rated sentence(s)}                                                                                    \\ \midrule
\textbf{True PQ Source}      & \cellcolor[HTML]{9AFF99}There is not a downtown in the whole wide world that’s made better by vehicle traffic.        \\ \midrule
\textbf{Quote\_count}       & We need to stop widening roads and otherwise “improving” our road infrastructure, and pronto.                         \\
\textbf{Sent\_position}     & By putting an immediate moratorium on it.                                                                             \\
\textbf{R\_difficult}        & But at the same time (this is the important part), make it super easy, free (or nearly free) and convenient to get    \\
\textbf{}                   & \quad around downtown.                                                                                                  \\
\textbf{POS\_PRP}           & Not, I think, if we have any say over it.                                                                             \\
\textbf{POS\_VB}            & Have them criss-cross the inner core.                                                                                 \\
\textbf{A\_concreteness}    & Not, I think, if we have any say over it.                                                                             \\ \midrule
\textbf{Char-2}             & We live far away from where we need to be, and we enjoy activities that aren’t always practical by bus, especially if \\
\textbf{}                   & \quad you happen to have kids that need to be in six different places every day.                                        \\
\textbf{Word-1}             & We live far away from where we need to be, and we enjoy activities that aren’t always practical by bus, especially if \\
\textbf{}                   & \quad you happen to have kids that need to be in six different places every day.                                        \\ \midrule
\textbf{C-deep}             & I want to scream.                                                                                                     \\ \midrule
\textbf{Headline popularity} & Personally, I’d rip out the Queensway and turn it into a light-rail line with huge bike paths, paths for motorcycles, \\
\textbf{}                   & \quad and maybe a lane or two dedicated to autonomous vehicles and taxis and ride-shares.                               \\
\textbf{Clickbait}          & It’s an idea I’ve been obsessed with since visiting Portland, Oregon, in 2004.                                        \\
\textbf{TextRank} & Not, I think, if we have any say over it.                                                                             \\ \bottomrule
\end{tabular}%
}
\caption{Article source: \url{https://ottawacitizen.com/opinion/columnists/armchair-mayor-fewer-cars-more-transit-options-would-invigorate-ottawa}}
\label{tab:pq_samples_3}
\end{table}

\begin{table}[h]
\centering
\resizebox{\textwidth}{!}{%
\begin{tabular}{@{}ll@{}}
\toprule
\textbf{Model}           & \textbf{Highest rated sentence(s)}                                                                      \\ \midrule
\textbf{True PQ Source} &
  \cellcolor[HTML]{9AFF99}But Pelosi seems to have thought more about alliteration than what pitch would effectively challenge the inaccurate but \\
\textbf{}                & \cellcolor[HTML]{9AFF99}\quad narratively satisfying story the president had just told.                   \\ \cmidrule(l){2-2} 
\textbf{} &
  \cellcolor[HTML]{9AFF99}Sanders packed more visceral humanity in the first minute or so of his remarks than in the entirety of Pelosi and \\
\textbf{}                & \cellcolor[HTML]{9AFF99}\quad Schumer’s response.                                                         \\ \cmidrule(l){2-2} 
\textbf{} &
  \cellcolor[HTML]{9AFF99}And perhaps most importantly, he validated that there is, in fact, a crisis afoot: one created by Trump, as well as \\
\textbf{}                & \cellcolor[HTML]{9AFF99}\quad several produced by structural forces the political class has long ignored. \\ \cmidrule(l){2-2} 
\textbf{} &
  \cellcolor[HTML]{9AFF99}And this is an important point: The temptation to fact-check is understandable. And a certain amount of fact-checking is \\
\textbf{} &
  \cellcolor[HTML]{9AFF99}\quad necessary to keep Trump accountable. But poking holes in Trump’s narrative, by itself, is not enough. \\ \midrule
\textbf{Quote\_count}    & The life of an American hero was stolen by someone who had no right to be in our country,” he said.     \\
\textbf{Sent\_position}  & An opioid crisis does kill thousands of Americans each year.                                            \\
\textbf{R\_difficult}    & The life of an American hero was stolen by someone who had no right to be in our country,” he said.     \\
\textbf{POS\_PRP}        & I’m not going to blame you {[}Chuck Schumer{]} for it.”                                                 \\
\textbf{POS\_VB} &
  I live paycheck to paycheck, and I can’t get a side job because I still have to go to my unpaid federal job.” \\
\textbf{A\_concreteness} & He didn’t disappoint.                                                                                   \\ \midrule
\textbf{Char-2}          & “Let me be as clear as I can be,” said Sanders, “this shutdown should never have happened.”             \\
\textbf{Word-1}          & “Let me be as clear as I can be,” said Sanders, “this shutdown should never have happened.”             \\ \midrule
\textbf{C-deep}          & All are equally guilty  -  children are merely “pawns,” not people.                                     \\ \midrule
\textbf{Headline popularity} &
  And what Trump said about who is hurting most is true: “Among the hardest hit are African-Americans and Hispanic- \\
\textbf{}                & \quad Americans.”                                                                                         \\
\textbf{Clickbait}       & “{[}Trump{]} talked about what happened the day after Christmas?                                        \\
\textbf{TextRank} &
  These are people in the FBI, in the TSA, in the State Department, in the Treasury Department, and other agencies who \\
\textbf{}                & \quad have, in some cases, worked for the government for years.”                                          \\ \bottomrule
\end{tabular}%
}
\caption{Article source: \url{https://theintercept.com/2019/01/09/trump-speech-democratic-response/}. This article demonstrates a case where there are many real PQs in an article. It also highlights the need for future work which can create multi-sentence PQs (True PQ \#4 consists of two sentences).}
\label{tab:pq_samples_4}
\end{table}

\end{appendices}
\newpage

\end{document}